\pdfoutput=1

\documentclass[11pt]{article}

\usepackage{acl}

\usepackage{times}
\usepackage{latexsym}

\usepackage[T1]{fontenc}

\usepackage[utf8]{inputenc}

\usepackage{microtype}

\usepackage{inconsolata}

\usepackage{graphicx}

\usepackage{inconsolata}
\usepackage{graphicx}

\usepackage{enumitem}
\usepackage{multirow}
\usepackage{booktabs}
\usepackage{algorithm}
\usepackage{algorithmic}
\usepackage{tikz}
\usepackage{pgfplots}
\usepackage{graphicx}
\usepackage{subcaption}
\usepackage{pgf-pie}
\usepackage{enumitem}
\usepackage{pifont}
\usepackage{amsmath}
\usepackage{supertabular}
\usepackage{booktabs}
\numberwithin{equation}{section}
\usepackage{dsfont}

\pgfplotsset{compat=1.18}
\usepackage{bm}
\usepackage{xspace}
\usepackage{tcolorbox}
\usepackage{bbding}
\usepackage{wasysym}
\usepackage{amssymb}
\usepackage{fontawesome5}

\newcommand{\ourmethod}{\textsc{CaR}\xspace}
\newcommand{\ourdataset}{\textsc{SciTaT}\xspace}
\definecolor{cpurple}{rgb}{0.675, 0.573, 0.922}
\definecolor{cblue}{rgb}{0.310, 0.757, 0.910}
\definecolor{cgreen}{rgb}{0.310, 0.563, 0.214}
\definecolor{corange}{rgb}{1, 0.808, 0.329}
\definecolor{cred}{rgb}{0.8, 0.3, 0.3}
\definecolor{data_blue}{rgb}{0.809, 0.883, 0.949}
\definecolor{annotator_pink}{rgb}{0.914, 0.816, 0.859}
\definecolor{reasoner_green}{rgb}{0.848, 0.914, 0.824}
\newcommand{\datatext}[1]{{\color{data_blue}#1}}
\newcommand{\annotatortext}[1]{{\color{annotator_pink}#1}}

\newtcolorbox{myexample}[2][]{
  colback=data_blue!40,
  colframe=data_blue,         
  coltitle=black,
  title=\textbf{#2},
  fonttitle=\bfseries,
  #1,
}

\title{\ourdataset: A Question Answering Benchmark for Scientific\\Tables and Text Covering Diverse Reasoning Types}

\author{
    Xuanliang Zhang$^1$\thanks{Correspondence to: \texttt{xuanliangzhang@ir.hit.edu.cn, car@ir.hit.edu.cn}}, Dingzirui Wang$^1$, Baoxin Wang$^2$, Longxu Dou$^3$,\\ 
    \textbf{Xinyuan Lu$^4$, Keyan Xu$^1$, Dayong Wu$^2$, Qingfu Zhu$^1$, Wanxiang Che$^{1*}$} \\
    $^1$Harbin Institute of Technology \quad $^2$iFLYTEK Research\\
    $^3$Individual Researcher \quad $^4$National University of Singapore
}

\begin{document}
    \maketitle
    \begin{abstract}
    Scientific question answering (SQA) is an important task aimed at answering questions based on papers. 
    However, current SQA datasets have limited reasoning types and neglect the relevance between tables and text, creating a significant gap with real scenarios.
    To address these challenges, we propose a QA benchmark for scientific tables and text with diverse reasoning types (\ourdataset).
    To cover more reasoning types, we summarize various reasoning types from real-world questions. 
    To involve both tables and text, we require the questions to incorporate tables and text as much as possible.
    Based on \ourdataset, we propose a strong baseline (\ourmethod), which combines various reasoning methods to address different reasoning types and process tables and text at the same time.
    \ourmethod brings average improvements of $12.9\%$ over other baselines on \ourdataset, validating its effectiveness.
    Error analysis reveals the challenges of \ourdataset, such as complex numerical calculations and domain knowledge.\footnote{Our codes and data are publicly available at \href{https://github.com/zhxlia/SciTaT}{https://github.com/zhxlia/SciTaT}.}

    \end{abstract}

    \section{Introduction}
Scientific Question Answering (SQA) plays a crucial role in addressing research questions based on scientific papers \cite{tsatsaronis2015overview-BIOASQ,pmlr-v202-QASA}.
Advancing SQA development can significantly accelerate knowledge acquisition \cite{taylor2022galactica,ai4science2023Scientific-Discovery}.
However, the dense technical terms and heterogeneous data representations in papers present challenges for the SQA task \cite{pramanick2024spiqa}.

To evaluate and enhance the model capabilities in SQA, numerous datasets are proposed \cite{pampari-etal-2018-emrqa,pappas-etal-2020-biomrc,jin-etal-2019-pubmedqa}.
However, existing datasets exhibit the following limitations, as shown in Table~\ref{tab:comparison_sciqa}. 
Firstly, the \textbf{reasoning types are relatively narrow}, failing to capture the complexity of real scenarios, such as data analysis, which is frequently encountered in actual queries.
Secondly, prior works \textbf{focus only on split text and tables} \cite{pmlr-v202-QASA,moosavi2021scigen,pramanick2024spiqa}, overlooking the relevance between tables and text, thereby limiting their applicability \cite{wang2022hybridqa-survey}.
To address the limitations, in this paper:
\textit{(i)} We introduce a new SQA benchmark, covering diverse real-scenery reasoning types and considering tables and text simultaneously.
\textit{(ii)} To enhance the performance on the benchmark, we propose a strong baseline, which is capable of handling multiple reasoning types and processing tables and text simultaneously.

\begin{figure*}
    \centering
    \includegraphics[width=.95\linewidth]{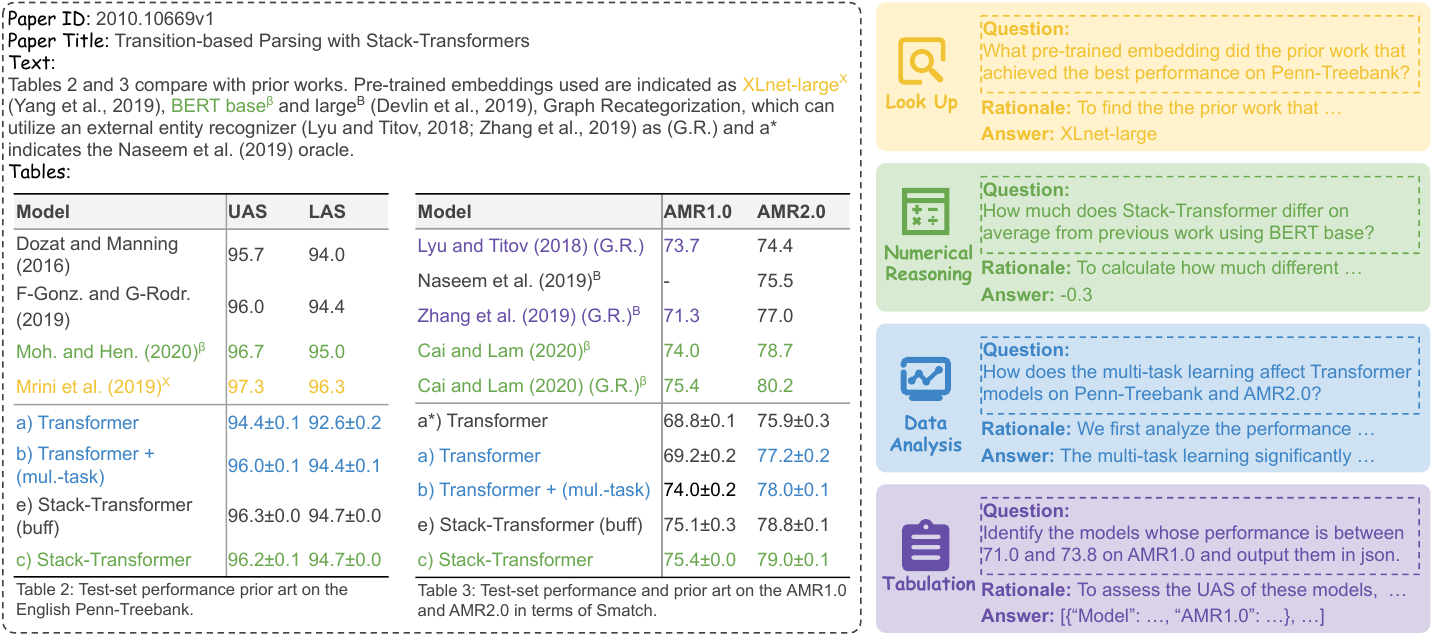}
    \vspace{-0.5em}
    \caption{
    Illustrations of the reasoning types in \ourdataset.
    The tables and text (left) show color-coded spans for question context.
    The questions (right) are examples of $4$ reasoning types, with their rationales and answers.
    }
    \label{fig:intro}
\end{figure*}

\begin{table}[t]
    \centering
    \small
    \begin{tabular}{l|cccc|ccc}
\toprule
\multirow{2}{*}{\textbf{Dataset}} & \multicolumn{4}{c}{\textbf{Reasoning Type}} & \multicolumn{3}{|c}{\textbf{Evidence}} \\
 & \textbf{L} & \textbf{N} & \textbf{D} & \textbf{T} & \textbf{Text} & \textbf{Table} & \textbf{TaT} \\
\midrule
BioRead  & \ding{51} & & & & \ding{51} & &\\
QASA  & \ding{51} & \ding{51} & \ding{51} &  & \ding{51} & &\\
SciGen & & & \ding{51} &  & & \ding{51} & \\
SciTab  & \ding{51} & \ding{51} & &  & & \ding{51} &\\
SPIQA   & \ding{51} & \ding{51} & &  & \ding{51} & \ding{51} &\\
\midrule
\ourdataset & \ding{51} & \ding{51} & \ding{51} & \ding{51} & \ding{51} & \ding{51} & \ding{51} \\
\bottomrule
\end{tabular}

    \vspace{-0.5em}
    \caption{
        Comparison of \ourdataset to recent SQA datasets.
        TaT denotes Table and Text.
        \textbf{L}, \textbf{N}, \textbf{D}, and \textbf{T} denote the reasoning type of Look Up, Numerical Reasoning, Data Analysis, and Tabulation, with examples in Figure~\ref{fig:intro}. 
        We introduce the datasets in Appendix~\ref{subsec:Comparison with Previous Scientific QA Datasets}.
    }
    \label{tab:comparison_sciqa}
\end{table}

Firstly, we propose a QA benchmark for scientific tables and text (\ourdataset), which are collected from papers in arXiv.org.
\textbf{To fit the actual scenario}, we summarize various reasoning types from the real questions raised by researcher (see Figure~\ref{fig:intro}).
\textbf{To ensure the questions involve both tables and text}, we request questions that involve both tables and text as much as possible.
Overall, \ourdataset contains $953$ questions derived from $871$ papers.
Data analysis reveals that \ourdataset encompasses $4$ reasoning types and $13$ subtypes, covering the types summarized from the real questions in SparkRA~\cite{wu-etal-2024-sparkra} and previous works \cite{lu-etal-2023-scitab,wu2024tablebench}.
\ourdataset not only requires the model to look up information and numerical reasoning but also requires complex data analysis and tabulation, effectively meeting the needs of researchers in real-world scenarios.

Considering the challenges of \ourdataset, we propose a strong baseline to process scientific data by integrating reasoning methods (\ourmethod).
\textbf{To handle multiple reasoning types}, \ourmethod includes two modules: Calculator and Reasoner. 
\textbf{To process tables and text}, Calculator extracts and calculates numerical information from tables and text, which is then provided to Reasoner for further reasoning.

We construct a series of baselines on \ourdataset. 
Experimental results reveal that \ourmethod outperforms other baselines with $12.9\%$ on average, proving the effectiveness of the combination of Calculator and Reasoner.
However, the Exact Match of \ourmethod using \texttt{gpt-4o} is still below $50\%$, which indicates that \ourdataset serves as a challenging benchmark.
Error analysis reveals the main challenges of \ourdataset, such as context grounding, complex numerical calculation, and the need for domain knowledge.

Our contributions are as follows:
\begin{enumerate}
    \item To the best of our knowledge, we develop \ourdataset, the first QA benchmark for scientific tables and text, covering diverse reasoning types based on real scenarios.
    \item We propose \ourmethod, a strong baseline to solve various reasoning types and process tables and text by integrating reasoning methods.
    \item We conduct a series of experiments, providing results and error analysis to highlight the challenges of \ourdataset, thereby guiding the direction for future improvements.
\end{enumerate}

    \section{\ourdataset Dataset}
        \label{sec:dataset}
\begin{figure*}
    \centering
    \includegraphics[width=.9\linewidth]{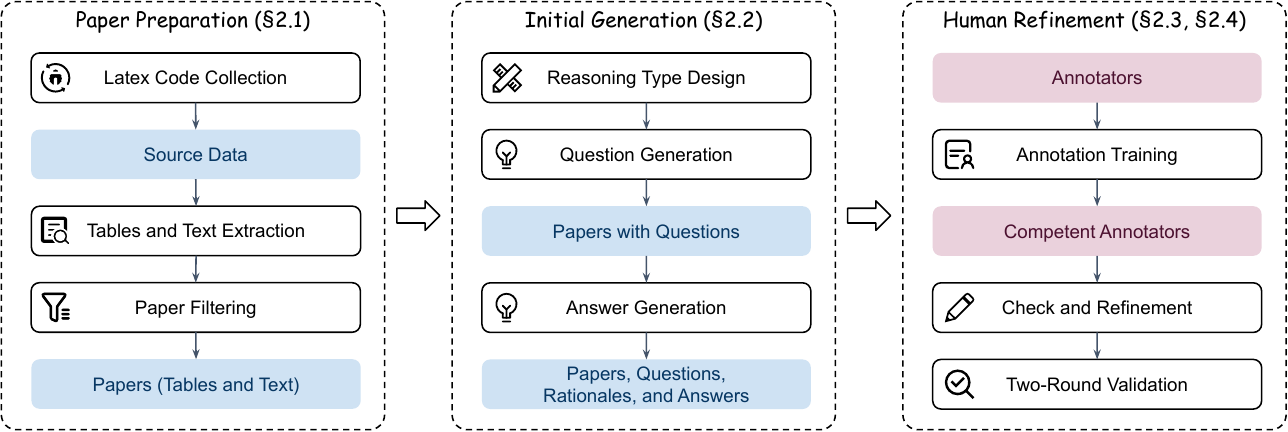}
    \vspace{-0.5em}
    \caption{
    The overview of our annotation process.
    The \datatext{blue} boxes represent the data, the \annotatortext{pink} boxes represent the annotators, and the white boxes with solid lines represent the annotation procedures.
    }
    \label{fig:annotation_process}
\end{figure*}

The input for the task associated with \ourdataset consists of scientific tables, text, and a question, and the output is the corresponding answer. 
Moreover, we annotate the rationale of each question.
For brevity, we refer to each question, its corresponding rationale, and answer, as an instance.
We begin by describing the construction process of \ourdataset.
We employ a framework combining automatic annotation with manual annotation to enhance both the quality and efficiency of the annotation process, as illustrated in Figure~\ref{fig:annotation_process}.

\subsection{Paper Preparation}
First, we introduce the data source and preparation of the scientific papers in \ourdataset.
We select papers from the ``Artificial Intelligence'', ``Computation and Language'', and ``Machine Learning'' subfields of ``Computer Science'' following previous datasets \cite{pmlr-v202-QASA,moosavi2021scigen,lu-etal-2023-scitab}.
We collect LaTeX code from papers published between January 2020 and July 2023 on arxiv.org, using a heuristic method to extract all the tables with their corresponding captions and labels, and text in each paper.
The tables are normalized into the \texttt{List[List[str]]} format.
To ensure the inclusion of both tables and text, we filter out papers without tables.
Additionally, to guarantee the relevance of the tables and text in the context, the context we provide when annotating the question is a paragraph that mentions tables and the tables mentioned.
Specifically, we randomly select at least one paragraph from the paper that mentions tables and the tables that are mentioned as the context of the question.

\subsection{Initial Generation}

\begin{figure*}[t]
\begin{myexample}{The requirements for generating questions}
1. The question must meet the reasoning type.\\
2. The question is best answered by referring to both the tables and the text simultaneously.\\
3. The question should be with fewer statements and more reasoning and calculation.
\end{myexample}
\vspace{-1.5em}
\caption{
The requirements for generating questions.
}
\label{fig:prompt_generate_question}
\end{figure*}

To observe the reasoning types that researchers might query, we select SParkRA~\cite{wu-etal-2024-sparkra}, a platform specifically designed to provide QA services for researchers in the context of scientific papers. 
We hypothesize that the reasoning types observed in these questions are comparable to those found in real-world inquiries. We randomly select $650$ questions and categorize their reasoning types. 
To account for potentially unobserved reasoning types, we also incorporate reasoning types from previous datasets \cite{lu-etal-2023-scitab, wu2024tablebench}.
Finally, we summarize $4$ reasoning types and $13$ subtypes for \ourdataset, as shown in Table~\ref{tab:reasoning_type}.

We assign a reasoning type to each context, including a paragraph and mentioned tables.
Manually annotating scientific questions and answers is time-consuming and prone to introducing annotation artifacts since it requires substantial domain expertise and a deep understanding of the paper \cite{bender2018data-statements,pramanick2024spiqa}.
To address these challenges, we leverage the extensive knowledge and powerful instruction-following capabilities of LLMs following previous works \cite{wu2024tablebench,zhang2024sciinstruct}. 
Specifically, we utilize \texttt{gpt-4o}~\cite{openai2024gpt4technicalreport} to generate questions and answers.
We guide the LLM to generate questions aligned with the reasoning type based on the context using three-shot prompt, with requirements outlined in Figure~\ref{fig:prompt_generate_question}.
We employ a Chain-of-Thought (CoT)~\cite{wei2022chain-of-thought} prompt with three demonstrations, guiding the LLM to generate the rationale and answer for each question according to the context. 
Detailed prompts are provided in Appendix~\ref{subsec:Prompt for Generating Data}.

\subsection{Human Refinement}
\label{subsec:Human Annotation}
Since LLMs cannot guarantee the reasonableness of the questions or the correctness of the answers, we employ manual checks and refinement.
(\emph{i}) For the context, annotators are tasked with verifying that the extracted tables and text are consistent with the original paper and removing any incorrect ones.
(\emph{ii}) For the questions, annotators should refine them following the guidelines in Figure~\ref{fig:prompt_generate_question}.
(\emph{iii}) For the rationales and answers, annotators are required to verify their correctness and correct any errors.
Due to the diverse reasoning types in \ourdataset, our answers include both \textit{short-form} and \textit{free-form} types. 
Annotators are instructed to extract one or more tokens for short-form answers and use complete sentences for free-form answers.
(\emph{iv}) For the answer source, annotators are prompted to select the source of the answer, which may include \textit{Text}, \textit{Table}, or \textit{Table and Text}, and identify the relevant tables.
Annotators are compensated \$1 per instance.

\subsection{Quality Control}
To ensure the quality of \ourdataset, we implement rigorous quality control strategies.

\subsubsection{Competent Annotators}
The annotators we employ are all graduate students majoring in artificial intelligence. 
Initially, the annotators undergo training sessions to learn the task and the annotation interface (see Appendix~\ref{subsec:Annotator Training Process}) and are required to annotate $20$ questions.
We retain those with Exact Match $\geq 95\%$ and provide constructive feedback on their mistakes.

\subsubsection{Two-round Validation}
After the instances are submitted by the annotator, a two-round validation process is implemented, consisting of manual verification and revision.
(\emph{i}) In the first round, a verifier examines each instance to ensure that the annotations adhere to the specified guidelines. If errors are found, the verifier communicates with the annotator and requests the corresponding corrections.
(\emph{ii}) In the second round, a different annotator reviews all the instances again. Any identified errors are discussed with the verifier and revised as needed.


\begin{table}[t]
    \centering
    \small


\begin{tabular}{l|cc}
\toprule
\textbf{Statistics} & \textbf{Long-context} & \textbf{Short-context} \\
\midrule
Questions & $953$ & $953$ \\
Papers & $871$ & $871$ \\
Avg. Tables & $5.2$ & $1.1$ \\
Avg. Cells & $60.8$ & $56.6$ \\
Avg. Paragraphs & $80.2$ & $1.0$ \\
Avg. |Paragraph| & $83.4$ & $113.0$ \\
\bottomrule
\end{tabular}

    \vspace{-0.5em}
    \caption{
        The statistics of \ourdataset.
        Avg. Tables and Avg. Cells indicate the average number of tables and the average number of cells per table.
        Avg. Paragraphs and Avg. |Paragraph| indicate the average number of paragraphs and the average length of each paragraph.
    }
    \label{tab:basic_statistics}
\end{table}

\begin{table}[t]
    \centering
    \small
    \begin{tabular}{l|cccc}
\toprule
\textbf{Statistics} & \textbf{Table} & \textbf{Text} & \textbf{TaT} & \textbf{Total}\\
\midrule
Short-form answers & $234$ & $13$ & $93$ & $340$ \\
Free-form answers & $308$ & $67$ & $238$ & $613$ \\
Total & $542$ & $80$ & $331$ & $953$ \\
\bottomrule
\end{tabular}
    \vspace{-0.5em}
    \caption{
        Question distribution over different answers and sources in \ourdataset. 
        TaT denotes the source of the Table and Text.
    }
    \label{tab:statistics_evidence}
\end{table}

\begin{table*}[ht]
\centering
\small
\begin{tabular}{lll c}
\toprule
\textbf{Reasoning Type} & \textbf{Subtypes} & \textbf{Description} & \textbf{\%} \\
\midrule
\multirow{2}{*}{Look Up ($4.7$)} 
  & Table Look Up & Search for specific tables & $2.7$ \\
  & Span Look Up & Search for spans in tables or paragraphs & $2.0$ \\
\midrule
\multirow{6}{*}{Numerical Reasoning ($46.0$)} 
  & Arithmetic Calculation & Numerical calculations & $11.1$ \\
  & Comparison & Comparison of values & $8.2$ \\
  & Aggregation & Combines multiple data points into a single metric & $3.9$ \\
  & Ranking & Arranges items in a specific order & $7.0$ \\
  & Counting & Counting occurrences & $9.2$ \\
  & Domain Knowledge Calculation & Calculations requiring domain knowledge & $6.5$ \\
\midrule
\multirow{3}{*}{Data Analysis ($40.9$)} 
  & Descriptive Analysis & Summarize or interpret to spot patterns and trends & $23.3$ \\
  & Anomaly Detection & Detect deviations and their causes & $7.0$ \\
  & Causal Analysis & Investigate cause-and-effect relationships & $10.6$ \\
\midrule
Tabulation ($8.4$) & - & Standardizing the formats of tables/subtables & $8.4$ \\
\bottomrule
\end{tabular}
\caption{
The reasoning types, the description of their subtypes, and their proportion in \ourdataset. 
The number in parentheses is the proportion of each reasoning type.
}
\label{tab:reasoning_type}
\end{table*}

\subsection{Data Analysis}

\subsubsection{Basic Statistics}
To better evaluate the reasoning ability across different context lengths, we propose two settings: \textit{long-context} and \textit{short-context}.
In the long-context setting, the model should answer questions based on the whole paper.
In the short-context setting, the model is required to answer questions based on a single paragraph and the tables referenced within that paragraph.
We present the statistics of \ourdataset in the long-context and short-context settings, as illustrated in Table~\ref{tab:basic_statistics}.
We also show the question distribution over different answers and sources in Table~\ref{tab:statistics_evidence}. 
Notably, over $1/3$ of the questions in \ourdataset require reasoning that involves both tables and text simultaneously.

\subsubsection{Reasoning Types}
We analyze the distribution of reasoning types in \ourdataset, as shown in Table~\ref{tab:reasoning_type}.
It can be found that \ourdataset has a variety of reasoning types evenly distributed.
Among these types, Data Analysis and Tabulation are identified as common patterns based on observations of real queries and are rarely represented in existing datasets.

    
    \section{\ourmethod}
        \label{sec:methodology}
        \begin{figure*}
    \centering
    \includegraphics[width=.9\linewidth]{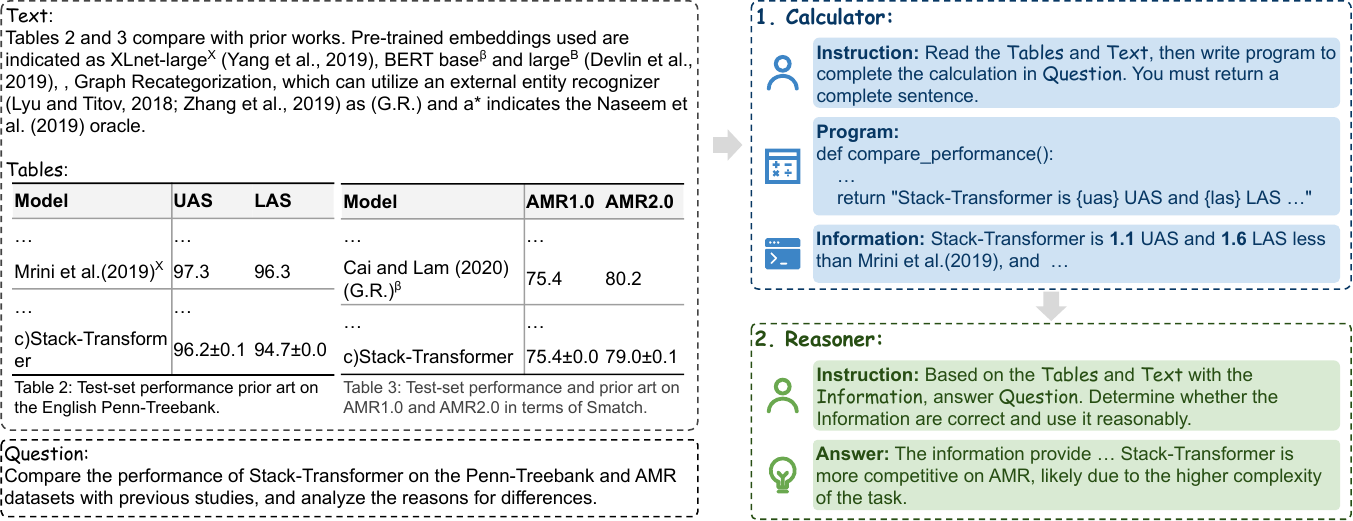}
    \caption{
    The overview of \ourmethod, which consists of two modules:
    (\emph{i})~Calculator generates code to compute the numerical information required for solving the question.
    (\emph{ii})~Reasoner continues the reasoning process based on the information provided by the Calculator to answer the question.
    }
    \label{fig:method}
\end{figure*}

\ourmethod is designed to address the questions on the context of scientific tables and text.
Given that \ourdataset combines diverse reasoning types, \ourmethod is composed into two modules: Calculator and Reasoner, as illustrated in Figure~\ref{fig:method}, which focus on different reasoning types.
To process tables and text simultaneously, the Calculator extracts and computes the numerical information from the context and the Reasoner can derive the final answer based on the calculated information.
The prompts we use are presented in Appendix~\ref{subsec:Prompt for Experiments}. 

\subsection{Calculator}
The input to the Calculator consists of a question and the scientific context (including tables and text), and the output provides the numerical information necessary to answer the question.
Specifically, we prompt the LLM to generate a program function based on the context to answer the question. 
Unlike other Program-of-Thought (PoT) methods \cite{pal,pot} that require the program to return the answer directly, the function is designed to return a sentence explicitly describing the numerical information, as illustrated in Figure~\ref{fig:method}.
Once the function is obtained, it is executed to extract the numerical information.

\subsection{Reasoner}
The Reasoner takes as input a question, the scientific context, and the numerical information obtained from the Calculator to produce the final answer.
Specifically, we utilize a CoT prompt \cite{wei2022chain-of-thought} to guide the LLM through a step-by-step reasoning process based on the context and information, leading to the final answer.
However, since the information may not always be accurate or helpful, we further prompt the LLM to engage in reflection, evaluating the correctness and relevance of the extracted information during reasoning.

    \section{Experiments}
        \label{sec:experiments}
        \subsection{Settings}
\subsubsection{Metrics}
Due to the significant difference in token counts between free-form answers and short-form answers (see Table~\ref{tab:token} in Appendix~\ref{subsec:Statistics of The Number of Output Tokens}), we evaluate the two types of answers separately.
For short-form answers, we use Exact Match (EM) to assess correctness, while for free-form answers, we use F1 and BERTScore F1 (BERTScore)~\cite{Zhang2020BERTScore} to evaluate accuracy, following previous studies \cite{zhu-etal-2021-tat,moosavi2021scigen}.
EM measures the proportion of the predicted result that exactly match the gold answer. 
F1 calculates the overlap between predicted and gold answers based on their bag-of-words representation.
BERTScore evaluates the similarity between generated text and reference text by calculating the cosine similarity of their word embeddings.

\subsubsection{Models}
We employ the open-source LLM Llama3.1-Instruct (Llama3.1)~\cite{dubey-etal-2024-llama3.1} and the closed-source LLM \texttt{gpt-4o}~\cite{openai2024gpt4technicalreport} to evaluate \ourdataset.
Llama3.1 is among the top-performing open-source models, while \texttt{gpt-4o} is one of the leading closed-source models.

\subsubsection{Baselines}
We compare \ourmethod with Direct QA~\cite{pramanick2024spiqa}, CoT~\cite{wei2022chain-of-thought}, and PoT~\cite{pal,pot} methods.
Direct QA refers to prompting the LLM to directly answer the questions.
The specific prompts for the baselines are provided in Appendix~\ref{subsec:Prompt for Experiments}. 
Given the long-context setting, where the context length is considerable, we adopt zero-shot prompts in experiments to prevent exceeding the context limit.



\begin{table*}[ht]
\centering
\small
\begin{tabular}{lll|ccc|ccc}
\toprule
\multirow{2}{*}{\textbf{Model}} & \multirow{2}{*}{\textbf{Scale}}  & \multirow{2}{*}{\textbf{Method}} & \multicolumn{3}{c|}{\textbf{Long-context}} & \multicolumn{3}{c}{\textbf{Short-context}} \\
 & & & \textbf{EM} & \textbf{F1} & \textbf{BERTScore} & \textbf{EM} & \textbf{F1} & \textbf{BERTScore} \\
\midrule
\multirow{10}{*}{Llama3.1} & \multirow{5}{*}{8B} & Direct QA & $0.0$ & $30.6$ & $66.4$ & $0.0$ & $30.7$ & $66.5$ \\
& & CoT & $13.0$ & $29.5$ & $65.6$ & $20.6$ & $41.4$ & $71.6$ \\
& & PoT & $4.4$ & $21.7$ & $54.0$ & $17.1$ & $21.5$ & $49.9$ \\
& & \ourmethod & \bm{$24.8$} & \bm{$37.5$} & \bm{$69.7$} & \bm{$24.2$} & \bm{$44.3$} & \bm{$73.2$} \\
& & $\Delta$ & $+19.1$ & $+10.2$ & $+7.7$ & $+11.6$ & $+13.1$ & $+10.5$ \\
\cmidrule{2-9}
& \multirow{5}{*}{70B} & Direct QA  & $0.0$ & $31.6$ & $67.5$ & $0.0$ & $33.6$ & $68.7$ \\
& & CoT & $30.3$ & $36.8$ & $69.9$ & $32.1$ & $44.1$ & $73.1$ \\
& & PoT & $5.3$ & $28.8$ & $61.7$ & $36.8$ & $35.6$ & $64.0$ \\
& & \ourmethod & \bm{$35.9$} & \bm{$41.7$} & \bm{$71.8$} & \bm{$40.7$} & \bm{$46.2$} & \bm{$74.4$} \\
& & $\Delta$ & $+24.2$ & $+9.3$ & $+5.4$ & $+17.7$ & $+8.4$ & $+5.8$ \\
\midrule
\multirow{5}{*}{\texttt{gpt-4o}} & \multirow{5}{*}{-} & Direct QA & $0.0$ & $29.8$ & $67.3$ & $0.0$ & $39.6$ & $71.5$ \\
& & CoT & $31.3$ & $41.3$ & $72.7$ & $32.2$ & $46.8$ & $75.5$ \\
& & PoT & $5.0$ & $15.0$ & $44.4$ & $28.3$ & $31.2$ & $59.8$ \\
& & \ourmethod & \bm{$37.5$} & \bm{$41.8$} & \bm{$73.1$} & \bm{$43.7$} & \bm{$47.1$} & \bm{$75.7$} \\
& & $\Delta$ & $+25.4$ & $+13.1$ & $+11.6$ & $+23.5$ & $+7.9$ & $+6.8$ \\
\bottomrule
\end{tabular}
\caption{
Performance comparison of different models and methods.
The best results of each model under each setting are annotated in \textbf{bold}. 
$\Delta$ refers to the average improvement of \ourmethod compared with other baselines.
}
\label{tab:main}
\end{table*}

\subsection{Main Experiments}
\label{subsec:main experiments}
The results of comparing \ourmethod with other baselines on \ourdataset are shown in Table~\ref{tab:main}. 
The results reveal that: 
(\emph{i})~\ourmethod significantly outperforms other baselines across different models and settings, achieving an average improvement of $12.9\%$ on all metrics, highlighting its effectiveness.
(\emph{ii})~Despite improvement, \ourmethod still demonstrates suboptimal performance, as both EM and F1 remain below $50.0$, and while BERTScore is relatively high \cite{moosavi2021scigen,zhao-etal-2024-tapera}, it remains under $80.0$, reflecting the challenge of \ourdataset.
We also observe that:

\paragraph{Baselines}
Among the three baselines, CoT achieves the highest overall performance, while Direct QA exhibits lower EM, and PoT shows lower F1 and BERTScore. 
Considering that a diverse range of reasoning types in \ourdataset, CoT is relatively better at handling these types of questions \cite{wei2022chain-of-thought,wu2024tablebench,pramanick2024spiqa}. 
Direct QA, due to its lack of reasoning, is prone to computational errors and incorrect grounding, and tends to generate longer answers for short-form answers, resulting in an EM score of zero \cite{snell2024scaling}. 
On the other hand, since the program typically returns shorter answers (see Appendix~\ref{subsec:Statistics of The Number of Output Tokens}), the PoT method is less effective at answering free-form questions.

\paragraph{Context Settings}
\ourmethod demonstrates a more significant improvement in the long-context setting than the short-context setting. 
Due to the dense knowledge presented in the paper, directly answering questions based on the entire paper may confuse the model, preventing it from focusing on the relevant tables and text \cite{pmlr-v202-QASA,pramanick2024spiqa}. 
In contrast, \ourmethod uses the Calculator to extract and compute useful numerical information from the paper, effectively guiding the Reasoner and avoiding the need to search for answers directly within the whole paper.

%

\paragraph{Answer Types}
\ourmethod shows more significant improvements in short-form answers than free-form answers.
For short-form answers, the Reasoner typically only needs to verify the correctness of the result of the Calculator and extract the answer. 
In contrast, for free-form answers, the Reasoner often needs to perform additional analysis based on the numerical information provided by the Calculator, which results in less significant improvement compared to short-form answers.


\begin{figure}[t]
    \centering
    \resizebox{0.75\linewidth}{!}{
    \begin{tikzpicture}
        \tiny
        \begin{axis}
            [ybar,
             xlabel=Reasoning Types,
             ylabel=Percentage,
             xmin=-0.1, xmax=3,
             ymin=0, ymax=100,
             width=\linewidth,
             xtick={0,1,2,3},
             xticklabels={Look Up,Numerical Reasoning,Data Analysis,Tabulation},
             legend style={
                at={(0.99,1.15)}, 
                anchor=north east, 
                font=\tiny, 
                legend cell align=left
             },
             every node near coord/.append style={
                font=\tiny,
             },
             bar width=7pt,
             xtick distance=12pt,
             axis y line=left,
             axis x line=bottom,
             nodes near coords,
             enlarge x limits=0.08, 
             nodes near coords={\pgfmathprintnumber[fixed,precision=2]{\pgfplotspointmeta}},
             ]
             
             \addplot+ [
                nodes near coords={\pgfmathprintnumber[fixed,precision=2]{\pgfplotspointmeta}},
                fill=reasoner_green!100,
                draw=reasoner_green!100, 
                every node near coord/.append style={text=reasoner_green!300} 
             ] 
             coordinates {
                (0,37.1)
                (1,37.0)
                (2,0)
                (3,15.4)
             }; 
             
             \addplot+ [
                ybar,
                fill=data_blue!100, 
                draw=data_blue!100, 
                nodes near coords={\pgfmathprintnumber[fixed,precision=2]{\pgfplotspointmeta}},
                every node near coord/.append style={text=data_blue!300} 
             ]
             coordinates {
                (0,0) 
                (1,48.0)
                (2,43.9)
                (3,52.5)
             };

            \addplot+ [
                ybar,
                fill=annotator_pink!100, 
                draw=annotator_pink!100, 
                nodes near coords={\pgfmathprintnumber[fixed,precision=2]{\pgfplotspointmeta}},
                every node near coord/.append style={text=annotator_pink!300} 
             ]
             coordinates {
                (0,0)
                (1,74.2)
                (2,72.9)
                (3,83.6)
             };
             
             \legend{EM, F1, BERTScore}; 
        \end{axis}
    \end{tikzpicture}
}
    \vspace{-0.5em}
    \caption{
        The average performance of \ourmethod across three models on four reasoning types.
    } 
    \label{fig:type_analysis}
    \vspace{-1em}
\end{figure}
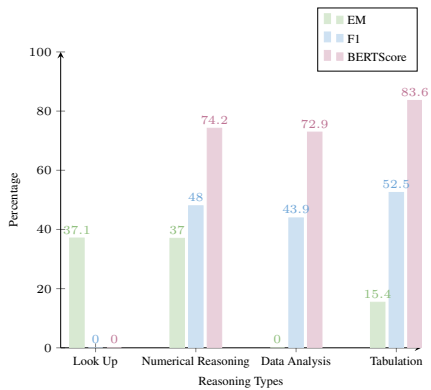

\paragraph{Reasoning Types}
We present the average performance of three models on different reasoning types in Figure~\ref{fig:type_analysis}. 
Specifically, the F1 and BERTScore for the type of Look Up are both $0$ as all questions in this type correspond to short-form answers. The EM for the type of Data Analysis is also $0$, as all the answers to this reasoning type are free-form. 
We can also observe that:
(\emph{i})~The models perform worst on Data Analysis, which requires more comprehensive capabilities, such as numerical computation, logical reasoning, summarization \cite{wu2024tablebench}. 
(\emph{ii})~The F1 and BERTScore on Tabulation are the highest, but the EM is the lowest, indicating the difficulty of this reasoning type. While the predicted result may be close to the gold answer, achieving an exact match remains challenging. This highlights the need for more effective evaluation metrics for this reasoning type.
(\emph{iii})~There is still significant room for improvement in the types of Look Up and Numerical Reasoning, underscoring the overall difficulty of \ourdataset.

\begin{figure}[t]
    \centering
    \resizebox{0.75\linewidth}{!}{
    \begin{tikzpicture}
        \tiny
        \begin{axis}
            [ybar,
             xlabel=Error Types,
             ylabel=Percentage,
             xmin=-0.1, xmax=4,
             ymin=0, ymax=25,
             width=\linewidth,
             xtick={0,1,2,3,4},
             xticklabels={Miss,Grounding,Calculation,Knowledge,Redundancy},
             legend style={
                at={(0.99,0.95)}, 
                anchor=north east, 
                font=\tiny, 
                legend cell align=left
             },
             every node near coord/.append style={
                font=\tiny,
             },
             bar width=7pt,
             xtick distance=12pt,
             axis y line=left,
             axis x line=bottom,
             nodes near coords,
             enlarge x limits=0.08, 
             nodes near coords={\pgfmathprintnumber[fixed,precision=2]{\pgfplotspointmeta}},
             ]
             
             \addplot+ [
                nodes near coords={\pgfmathprintnumber[fixed,precision=2]{\pgfplotspointmeta}},
                fill=data_blue!100,
                draw=data_blue!100,
                every node near coord/.append style={text=data_blue!300} 
             ] 
             coordinates {
                (0,24)
                (1,18)
                (2,4)
                (3,4)
                (4,0)
             }; 
             
             \addplot+ [
                ybar,
                fill=annotator_pink!100, 
                draw=annotator_pink!100,
                nodes near coords={\pgfmathprintnumber[fixed,precision=2]{\pgfplotspointmeta}},
                every node near coord/.append style={text=annotator_pink!300} 
             ]
             coordinates {
                (0,6) 
                (1,6)
                (2,20)
                (3,4)
                (4,14)
             };
             
             \legend{Free-form answers, Short-form answers}; 
        \end{axis}
    \end{tikzpicture}
}
    \caption{
        The distribution of error types of \ourmethod on free-form answers and short-form answers.
    } 
    \label{fig:error}
\end{figure}
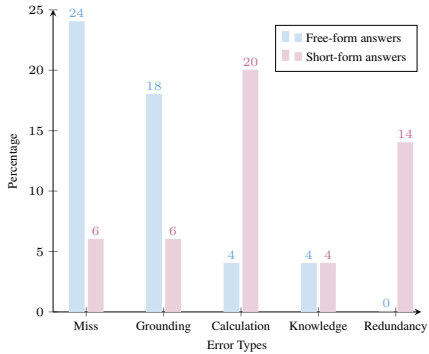

\begin{table}[t]
\centering
\small
\begin{tabular}{ll|ccc}
\toprule
\textbf{Scale} & \textbf{Method} & \textbf{EM} & \textbf{F1} & \textbf{BERTScore} \\ 
\midrule
\multirow{4}{*}{8B} & \ourmethod & \bm{$24.2$} & \bm{$44.3$} & \bm{$73.2$} \\ 
 & \textit{Reasoner} & $20.6$ & $41.4$ & $71.6$ \\ 
 & \textit{Calculator} & $0.3$ & $30.8$ & $62.8$ \\ 
 & \textit{Reversing} & $21.5$ & $21.4$ & $47.2$ \\ 
\midrule
\multirow{4}{*}{70B} & \ourmethod & \bm{$40.7$} & \bm{$46.2$} & \bm{$74.4$} \\ 
 & \textit{Reasoner} & $32.1$ & $44.1$ & $73.1$ \\ 
 & \textit{Calculator} & $0.0$ & $42.0$ & $70.4$ \\ 
 & \textit{Reversing} & $37.1$ & $35.9$ & $65.1$ \\ 
\bottomrule
\end{tabular}
\caption{
The ablation results of \ourmethod, compared with only Reasoner, only Calculator, and reversing the two modules (denoted as \textit{Reversing}) on \ourdataset on the short-context setting. 
}
\label{tab:ablation}
\end{table}

\subsection{Ablation Experiments}
To demonstrate the effectiveness of \ourmethod, we perform an ablation study by removing each module and reversing the order of the two modules. 
Specifically, when only the Reasoner is retained, it is the same as the CoT baseline. 
When reversing, we first apply the Reasoner module and then feed its output into the Calculator, which verifies and corrects any numerical errors to produce the final result. 
We perform experiments using Llama3.1 in the short-context setting, with results presented in Table~\ref{tab:ablation}. 
The significant performance drop confirms the validity of \ourmethod. 
The results also suggest that: 
(\emph{i})~Given the diverse reasoning types in \ourdataset, relying on a single reasoning method is insufficient to derive accurate answers. 
Especially, the EM of the Calculator is very low since we prompt the program to return the entire numerical information instead of the simple answer.
(\emph{ii})~Furthermore, as discussed in \S\ref{subsec:main experiments}, the program struggles with free-form responses, so depending solely on its output for the final answer limits performance on \ourdataset.

\subsection{Error Analysis}
In this section, we analyze the erroneous data of \ourmethod using Llama3.1-70B. 
Specifically, we randomly select $25$ error instances with BERTScore below $60$ from the results corresponding to free-form answers and another $25$ instances with EM of $0$ from the results corresponding to short-form answers. 
We manually categorize the error types, as illustrated in Figure~\ref{fig:error}. 
It can be observed that the distribution of error types for free-form answers and short-form answers differs significantly. 
Examples of different error types are provided in Appendix~\ref{sec:error case}. 
We proceed with a detailed analysis.

(\emph{i})~Miss refers to the omission of part of the answer, such as when only some sub-questions are addressed in a multi-part question, or when data analysis is limited to summarizing phenomena without providing conclusions or insights.
(\emph{ii})~Grounding refers to locating incorrect relevant context according to the question.
(\emph{iii})~Calculation denotes errors in applying numerical formulas, coding mistakes, or computational inaccuracies in complex calculations.
(\emph{iv})~Knowledge refers to errors in responses due to the lack of domain-specific knowledge.
(\emph{v})~Redundancy refers to the generation of unnecessary responses that result in an EM of zero.

The reasoning types for both free-form and short-form answers in \ourdataset exhibit diversity and are not entirely identical. 
Compared to previous datasets, \ourdataset presents the following main challenges:
(\emph{i})~Different error types are associated with free-form and short-form answers, which require the design of distinct methods for each.
(\emph{ii})~The need to integrate various reasoning types and process both tables and text, while requiring the model to possess strong domain-specific knowledge in the scientific domain.
We outline these challenges to inspire future work in developing methods that address these issues, aiming to enhance model performance in scientific QA on tables and text.

    \section{Related Works}
        \label{sec:related}
        \subsection{Scientific QA Datasets}
Early SQA datasets were designed in a cloze-style format, which limited the difficulty of these datasets \cite{pampari-etal-2018-emrqa,pappas-etal-2018-bioread}. 
To address this issue, PubMedQA~\cite{jin-etal-2019-pubmedqa}, QASPER~\cite{dasigi-etal-2021-QASPER}, and QASA~\cite{pmlr-v202-QASA} employ humans to annotate questions and answers over papers, and SciInstruct~\cite{zhang2024sciinstruct} collects questions from sources like textbooks and synthesizes answers using LLMs. 
However, these works primarily focus on text, without considering the tables often appearing in papers. 
Therefore, SciGen~\cite{moosavi2021scigen} focuses on generating relevant descriptions based on tables in papers, SciTab~\cite{lu-etal-2023-scitab} concentrates on the table fact verification task, and SPIQA~\cite{pramanick2024spiqa} is designed for QA based on tables and images.

Nevertheless, the reasoning types of existing datasets are relatively limited, since they do not involve diverse reasoning types, such as Data Analysis and Tabulation, that frequently occur in real scenarios.
Moreover, they overlook the relevance between tables and text, limiting their application \cite{chen-etal-2020-hybridqa,wang2022hybridqa-survey}. 
Therefore, we propose \ourdataset, a QA benchmark for scientific tables and text with diverse reasoning types. 


\subsection{QA Datasets for Tables and Text}
Previous QA datasets for tables and text mainly focus on Look Up and Numerical Reasoning in the Wikipedia and financial domains. 
For example, HybridQA~\cite{chen-etal-2020-hybridqa} annotates QA pairs over Wikipedia tables and text, which primarily focuses on look up spans in the context. 
TAT-QA~\cite{zhu-etal-2021-tat}, FinQA~\cite{chen-etal-2021-finqa}, and DocMath-Eval~\cite{zhao-etal-2024-docmath} primarily address the numerical reasoning task in the financial domain.
Among them, DocMath-Eval~\cite{zhao-etal-2024-docmath} collects data from previous datasets and annotates the Python program for each question. 
However, their reasoning types are relatively limited, which significantly differs from the scientific QA scenarios in real-world applications. 
A comparison \ourdataset with previous datasets for tables and text is shown in Table~\ref{tab:comparison_tat} of Appendix~\ref{subsec:Comparison with Previous QA Datasets for Tables and Text}.

Considering the reasoning types of existing datasets, previous works introduce programs to obtain the final answer \cite{pal,pot}.
For instance, TAT-LLM~\cite{TAT-LLM} proposes to first extract relevant context and generate equations or logical reasoning steps to execute them and derive the answer. 
Hpropro~\cite{shi-etal-2024-hpropro} provides commonly used program functions, allowing the LLM to directly call them during program generation.
However, these methods can not apply directly to \ourdataset, as \ourdataset also involves reasoning types, such as Data Analysis, which requires free-form answers,  challenging to be solved by the program alone \cite{wu2024tablebench}. 
Therefore, we propose \ourmethod, which combines multiple reasoning types to enhance performance on \ourdataset.
    
    \section{Conclusion}
    To address the limitations of previous scientific QA datasets, which involve limited reasoning types and fail to consider the relevance between tables and text, we propose \ourdataset, the QA benchmark for scientific tables and text with diverse reasoning types. 
    To incorporate diverse reasoning types, we analyze the questions posed by researchers and combine the types in prior works, summarizing $4$ reasoning types with $13$ subtypes. 
    To ensure that the questions encompass both tables and text, we require the questions include both elements whenever possible. 
    For \ourdataset, we introduce \ourmethod, a strong baseline that combines reasoning methods to enhance the performance across various reasoning types, with handling both tables and text. 
    Experimental results show that \ourmethod outperforms other baselines by an average of $12.9\%$, demonstrating its effectiveness. 
    Error analysis reveals that the challenges in \ourdataset, such as  grounding relevant context, complex numerical reasoning, and the need for domain-specific knowledge.

    
    \section*{Limitations}
        (\emph{i})~\ourdataset currently supports only the English language. Future versions will include additional languages.
        (\emph{ii})~Currently, we focus on single-turn QA for scientific tables and texts in \ourdataset. Multi-turn dialogues on scientific tables and text will be explored in future work.
        
    \section*{Ethics Statement}
            
    All datasets and models used in this paper are publicly available, and our utilization of them strictly complies with their respective licenses and terms of use.
    
    \bibliography{custom}
    
    \clearpage
    \appendix
    \label{sec:appendix}
    \section{Comparison with Previous Datasets}
\label{sec:comparison}

\subsection{Comparison with Previous Scientific QA Datasets}
\label{subsec:Comparison with Previous Scientific QA Datasets}
Table~\ref{tab:comparison_sciqa} presents the comparison of \ourdataset with previous scientific QA datasets.
We first introduce the existing datasets.
BioRead~\cite{pappas-etal-2018-bioread} is a cloze-style QA dataset on the biomedical papers, which only conteains the reasoning type of Look Up and focuses only on the text. 
QASA~\cite{pmlr-v202-QASA} is QA datasets on papers in AI and ML fields, but only concentrates on the text in papers, and lack the reasoning type of Tabulation.
SciGen~\cite{moosavi2021scigen} aims to generate descriptions according to the tables in the papers in the field of Computer Science.
SciTab~\cite{lu-etal-2023-scitab} aims to judge the claims according to the scientific tables in the field of Computer Science, which only contains the reasoning types of Lok Up and Numerical Reasoning.
SPIQA~\cite{pramanick2024spiqa} is mulmimodal QA dataset on the scientific papers, which only focus on the split text and tables, ignoring the relevance between tables and text, and lacking the reasoning type of Data Analysis and Tabulation.
It can be seen that \ourdataset contains more diverse reasoning types and consider the relevance between tables and text.

\subsection{Comparison with Previous QA Datasets for Tables and Text}
\label{subsec:Comparison with Previous QA Datasets for Tables and Text}
Table~\ref{tab:comparison_tat} present the comparison of \ourdataset with previous QA datasets over tabular and textual data and scientific QA datasets.
It can be seen that \ourdataset contains more diverse and closer to real-life user questions.

\begin{table*}[ht]
    \centering
    \small
    \begin{tabular}{l|lccccc}
\toprule
\multirow{2}{*}{\textbf{Dataset}} & \multirow{2}{*}{\textbf{Domain}} & \multicolumn{4}{c}{\textbf{Reasoning Type}} & \multirow{2}{*}{\textbf{R}} \\ 
\cmidrule{3-6} 
 & & \textbf{Look Up} & \textbf{Numerical Reasoning} & \textbf{Data Analysis} & \textbf{Tabulation} & \\ 
 \midrule
HybridQA~\cite{chen-etal-2020-hybridqa} & Wiki & \ding{51} & \ding{55} & \ding{55} & \ding{55} & \ding{55} \\
TAT-QA~\cite{zhu-etal-2021-tat} & Finance & \ding{51} & \ding{51} & \ding{55} & \ding{55} & \ding{55} \\ 
FinQA~\cite{chen-etal-2021-finqa} & Finance & \ding{55} & \ding{51} & \ding{55} & \ding{55} & \ding{55} \\ 
DocMath-Eval~\cite{zhao-etal-2024-docmath} & Finance & \ding{51} & \ding{51} & \ding{55} & \ding{55} & \ding{51} \\
\ourdataset & Science & \ding{51} & \ding{51} & \ding{51} & \ding{51} & \ding{51} \\ 
\bottomrule
\end{tabular}
    \caption{
        Comparison of \ourdataset to recent QA datasets over tabular and textual data.
        Wiki denotes Wikipedia, and R denotes Rationale.
    }
    \label{tab:comparison_tat}
\end{table*}

\section{Prompt}
In this section, we show the prompts we use to synthesize data and conduct experiments.

\subsection{Prompt for Generating Data}
\label{subsec:Prompt for Generating Data}
Table~\ref{tab:prompt_data} provides the prompt for generating questions, rationales, and answers when constructing \ourdataset.

\begin{table*}[ht]
    \centering
    \small
    \begin{tabular}{p{0.9\textwidth}}
\toprule
\textbf{The prompt for Generating Questions} \\
\midrule
\{Table\}\\
\{Paragraph\}\\
You are a highly intelligent and obedient academic field question generation system.\\
Generate a question referring to the table and paragraph above which meets the requirements in the question description "\{Type\}". \\
The generated question must meet:\\
1. The question should be with fewer statements and more reasoning and calculation.\\
2. The question must be answerable based on the paragraph alone, and not answerable only based on the table.\\
3. The question must meet the question description.\\
4. Do not generate multiple questions or sub-questions at once.\\
Examples:\\
\{Examples\}\\
\bottomrule
\end{tabular}

\begin{tabular}{p{0.9\textwidth}}
\toprule
\textbf{The prompt for Generating Rationales and Answers} \\
\midrule
\{Table\}\\
\{Paragraph\}\\
Based on the information in the Table and Paragraph, please answer the question "\{Question\}".\\
Represent your answer with: "Reason: <Your Reason> Answer: <Your Answer>"\\
If there are multiple questions, you need to answer them one by one, and the answers are separated by "\\ \\".\\
Examples:\\
\{Examples\}\\
\bottomrule
\end{tabular}
    \caption{
    The prompts for generating the questions, rationales, and answers of \ourdataset.
    }
    \label{tab:prompt_data}
\end{table*}

\subsection{Prompt for Experiments}
\label{subsec:Prompt for Experiments}
Table~\ref{tab:prompt_baseline} shows the prompt to build the baselines in our experiments, and Table~\ref{tab:prompt_our_method} shows the prompt used by \ourmethod.

\begin{table*}[ht]
    \centering
    \small
    \begin{tabular}{p{0.9\textwidth}}
\toprule
\textbf{The prompt for DirectQA} \\
\midrule
Based on the information in the Table and Paragraph, you should answer the question.\\
If there are multiple questions, you need to answer them one by one, and the answers are separated by "\\ \\".\\

Table (including its label, caption, and content):\\
\{Table\}\\
Paragraph:\\
\{Paragraph\}\\
\\
Please answer the question "\{Question\}".\\
\bottomrule
\end{tabular}

\begin{tabular}{p{0.9\textwidth}}
\toprule
\textbf{The prompt for CoT} \\
\midrule
Based on the information in the Table and Paragraph, you should answer the question.\\
Represent your answer with: "Reason: <Your Reason> Answer: <Your Answer>".\\
If there are multiple questions, you need to answer them one by one, and the answers are separated by "\\ \\".\\
\\
Table (including its label, caption and content):\\
\{Table\}\\
Paragraph:\\
\{Paragraph\}\\
\\
Please answer the question "\{Question\}".\\
\bottomrule
\end{tabular}

\begin{tabular}{p{0.9\textwidth}}
\toprule
\textbf{The prompt for PoT} \\
\midrule
Table (including its label, caption and content):\\
\{Table\}\\
Paragraph:\\
\{Paragraph\}\\
Read the above Table and Paragraph, and then write code to answer the question "\{Question\}".\\
Please **directly use** the information such as numbers in tables and paragraphs, do not define tables and then process them.\\
You must return the answer `ans = ` at the end of the code instead of `print`.\\
Attention that if there are multiple questions, you need to answer them one by one, and the answers are separated by "\\ \\".\\
\bottomrule
\end{tabular}
    \caption{
    The prompts for baselines.
    }
    \label{tab:prompt_baseline}
\end{table*}

\begin{table*}[ht]
    \centering
    \small
    \begin{tabular}{p{0.9\textwidth}}
\toprule
\textbf{The prompt for Calculator} \\
\midrule
Table (including its label, caption, and content):\\
\{Table\}\\
Paragraph:\\
\{Paragraph\}\\
Read the above Table and Paragraph, and then write code to answer the question "\{Question\}".\\
Please **directly use** the information such as numbers in tables and paragraphs, do not define tables and then process them.\\
You must return the answer `ans = ` at the end of the code instead of `print`.\\
You cannot return just one or a few numbers or words, you must return a complete sentence.\\
\bottomrule
\end{tabular}

\begin{tabular}{p{0.9\textwidth}}
\toprule
\textbf{The prompt for Reasoner} \\
\midrule
Based on the Table and Paragraph with the Tips, you should answer the question.\\
Please determine whether the tips are correct, use the tips reasonably in Reason, and organize the Answer into an appropriate form.\\
Represent your answer with: "Reason: <Your Reason> Answer: <Your Answer>".\\
Attention that if there are multiple questions, you need to answer them one by one, and the answers are separated by "\\ \\".\\
\\
Table (including its label, caption, and content):\\
\{Table\}\\
Paragraph:\\
\{Paragraph\}\\
Tips:\\
\{Tips\}
\\
Please answer the question "\{Question\}".\\
\bottomrule
\end{tabular}
    \caption{
    The prompts for \ourmethod.
    }
    \label{tab:prompt_our_method}
\end{table*}

\section{Manual Annotation Procedure}
\label{sec:Manual Annotation Procedure}

\subsection{Annotator Training Process}
\label{subsec:Annotator Training Process}
We recruit students from Computer Science or Artificial Intelligence programs who are willing to participate in the annotation task, offering a compensation of $\$1$ per instance. 
Initially, we provide a detailed explanation of the task, including its definition, the specific responsibilities of the annotators, and how to use the annotation interface. 
We thoroughly explain the requirements for the questions, rationales, and answers, as well as how to select the source of the answers, as stated in \S\ref{subsec:Human Annotation}.
Additionally, we provide three examples and explain possible scenarios that might arise. 
Finally, we clarify the annotation deadline and inform them that the data will undergo additional checks. 
To promptly detect any errors or biases in the annotations, we sent the data in batches. 
After the two-round validation on the already annotated data, we communicate with the annotators to address any issues and proceed to send the next batch of data.

\subsection{Statistics of the Manual Annotation Procedure}
On average, annotating a single data point required $10$ minutes per annotator. The annotation process for the $953$ instances was completed in approximately two months.
The first round of annotations was conducted by $10$ annotators, with two additional annotators performing two-round validation. 

\subsection{Annotating Interface}
\begin{figure*}[t]
    \centering
    \includegraphics[width=1\linewidth]{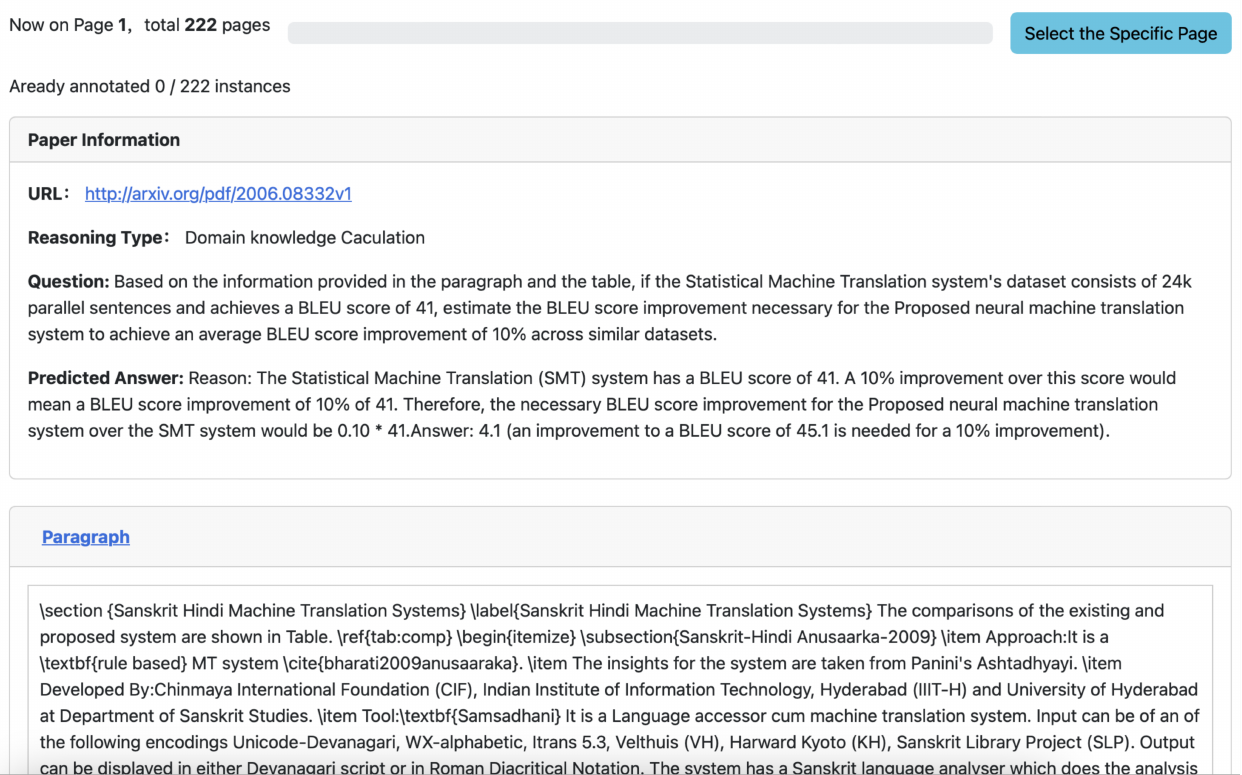}
    \caption{
    The user interface, showing the paper information and the paragraph.
    }
    \label{fig:interface1}
\end{figure*}

\begin{figure*}[ht]
    \centering
    \includegraphics[width=1\linewidth]{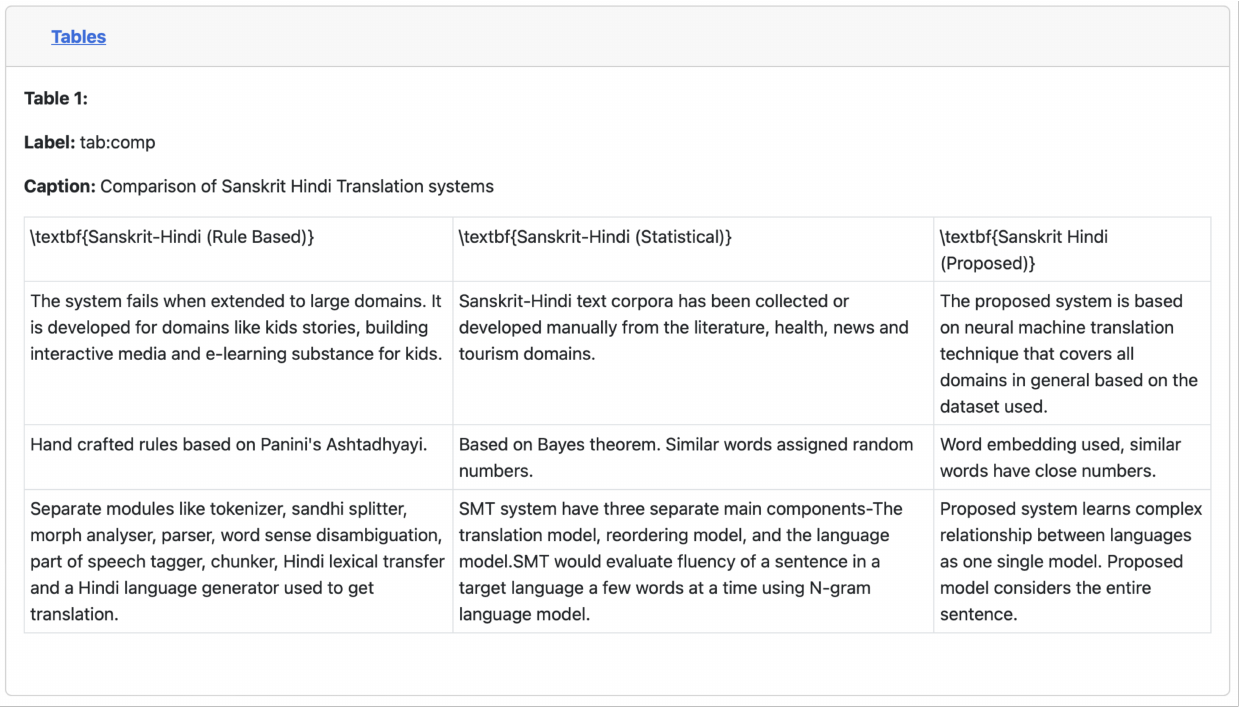}
    \caption{
    The user interface, showing the tables.
    }
    \label{fig:interface2}
\end{figure*}

\begin{figure*}[t]
    \centering
    \includegraphics[width=1\linewidth]{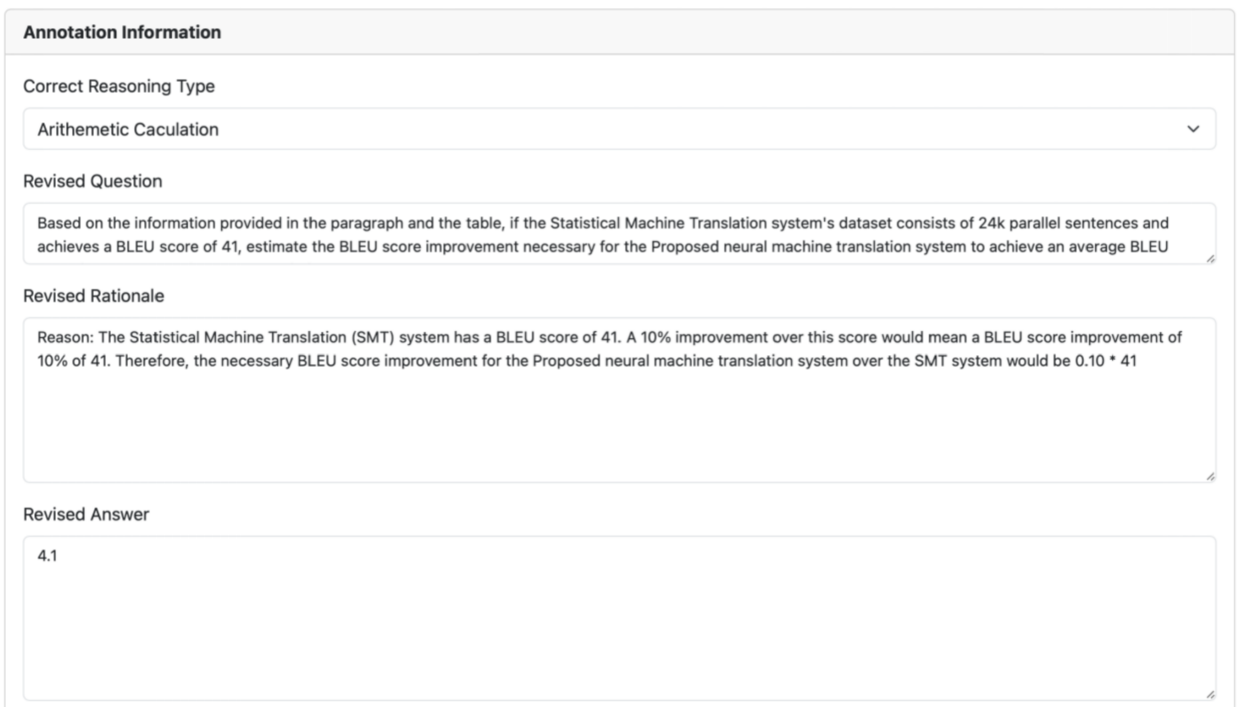}
    \caption{
    The user interface, showing the annotation information.
    }
    \label{fig:interface3}
\end{figure*}

\begin{figure*}[ht]
    \centering
    \includegraphics[width=1\linewidth]{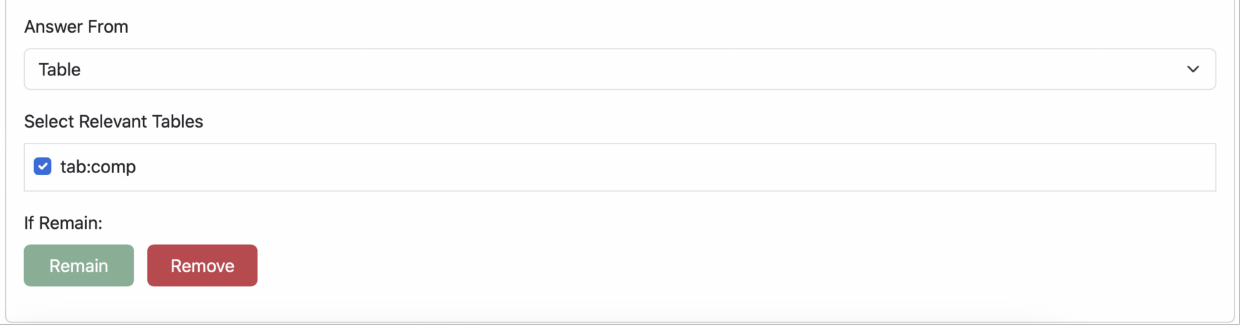}
    \caption{
    The user interface, showing the choice for the answer source, relevant tables, and if to remain.
    }
    \label{fig:interface4}
\end{figure*}

The annotation process is conducted using a custom tool developed by us.
Figure~\ref{fig:interface1}, Figure~\ref{fig:interface2}, Figure~\ref{fig:interface3}, and Figure~\ref{fig:interface4}  show the overall user interface for the manual annotation.

\section{Case Study for Error Analysis}
\label{sec:error case}
\begin{figure*}
    \centering
    \includegraphics[width=1\linewidth]{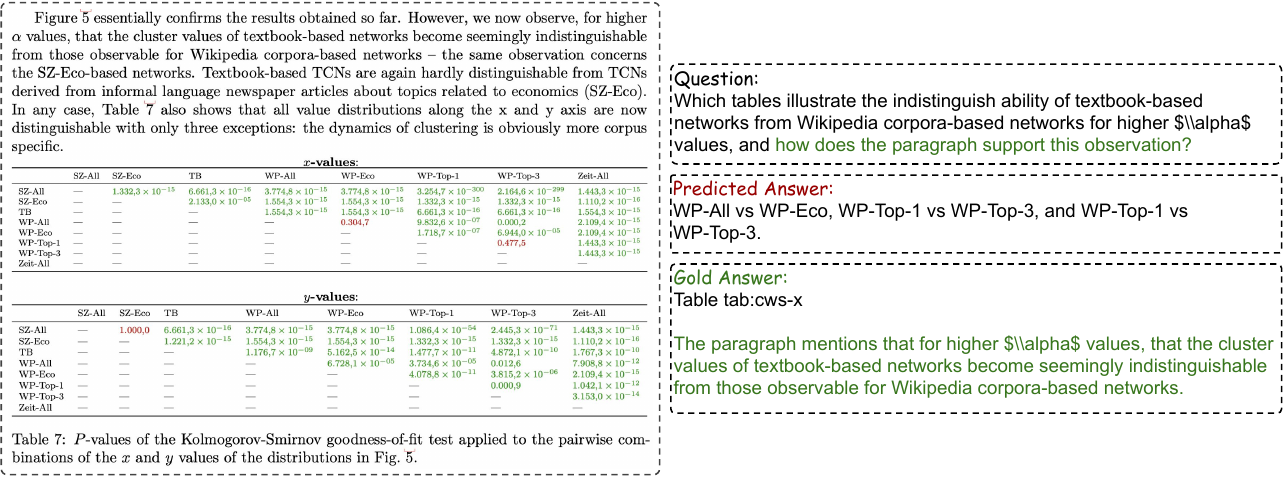}
    \caption{
    The case for the error type of "Miss".
    }
    \label{fig:error_miss}
\end{figure*}

\begin{figure*}
    \centering
    \includegraphics[width=1\linewidth]{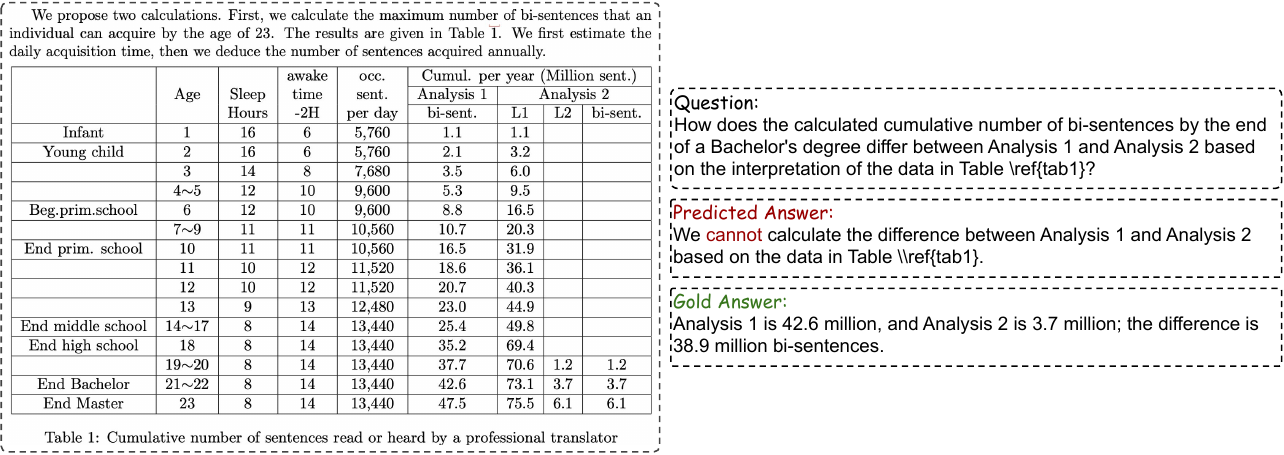}
    \caption{
    The case for the error type of "Grounding".
    }
    \label{fig:error_gounding}
\end{figure*}

\begin{figure*}
    \centering
    \includegraphics[width=1\linewidth]{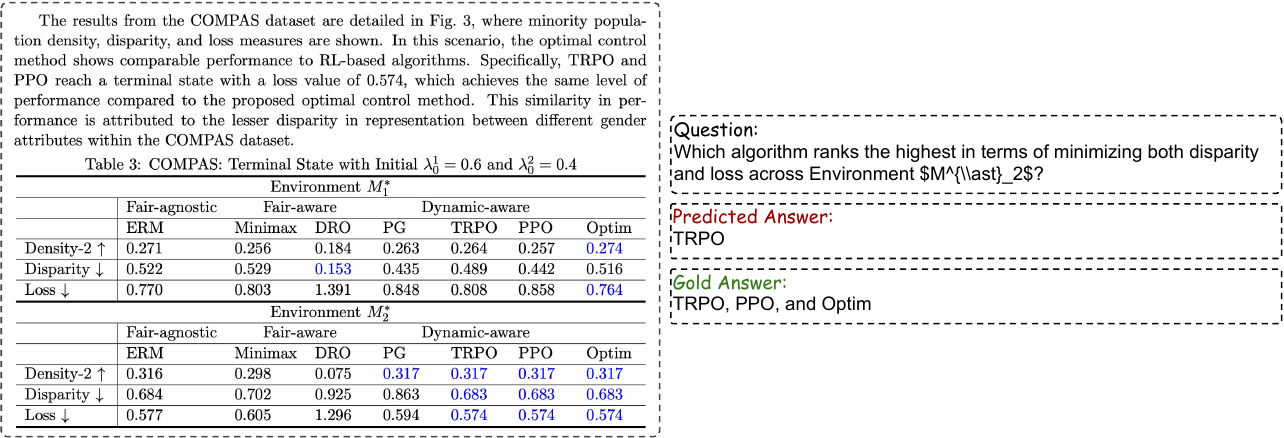}
    \caption{
    The case for the error type of "Calculation".
    }
    \label{fig:error_calculation}
\end{figure*}

\begin{figure*}
    \centering
    \includegraphics[width=1\linewidth]{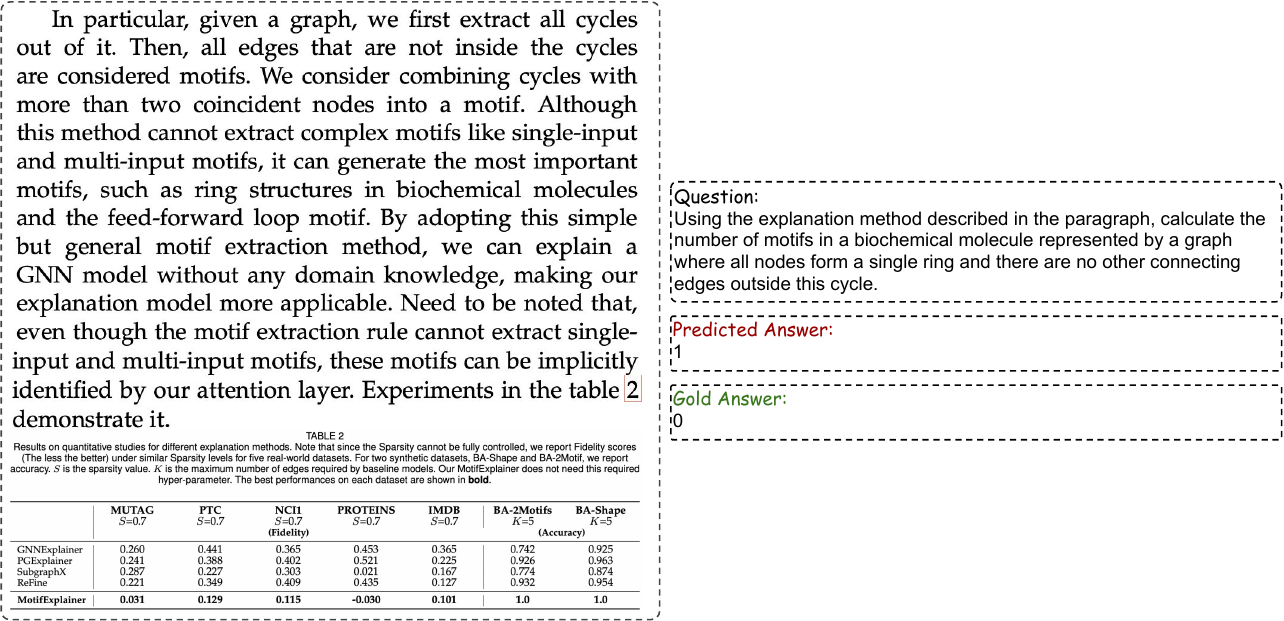}
    \caption{
    The case for the error type of "Knowledge".
    }
    \label{fig:error_knowledge}
\end{figure*}

\begin{figure*}
    \centering
    \includegraphics[width=1\linewidth]{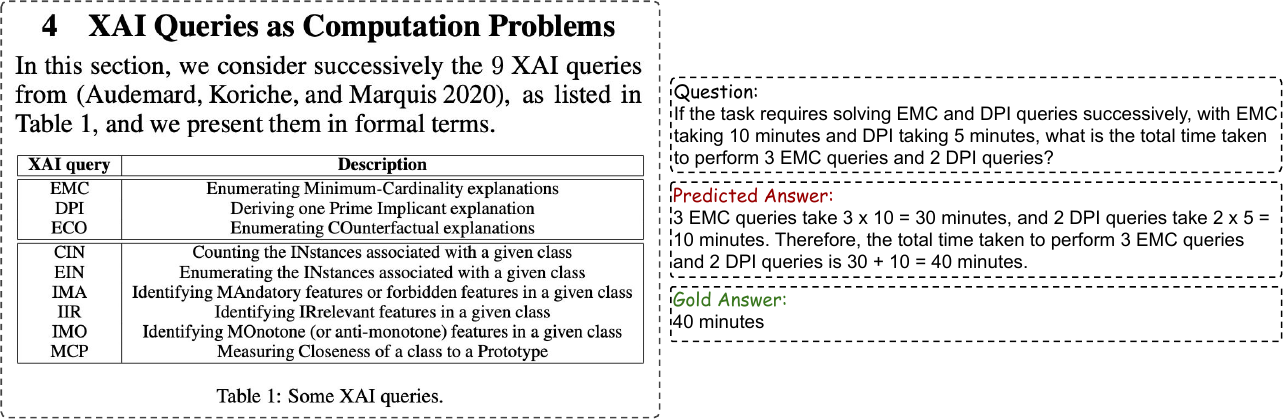}
    \caption{
    The case for the error type of "Redundancy".
    }
    \label{fig:error_redundancy}
\end{figure*}

In this section, we show examples of different error types, as shown in Figure~\ref{fig:error_miss}, Figure~\ref{fig:error_gounding}, Figure~\ref{fig:error_calculation}, Figure~\ref{fig:error_knowledge}, and Figure~\ref{fig:error_redundancy}.

\section{Additional Experiments}

\subsection{Statistics of The Number of Output Tokens}
\label{subsec:Statistics of The Number of Output Tokens}

\begin{table*}[ht]
\centering
\small
\begin{tabular}{lll|cc|cc}
\toprule
\multirow{2}{*}{\textbf{Model}} & \multirow{2}{*}{\textbf{Scale}}  & \multirow{2}{*}{\textbf{Method}} & \multicolumn{2}{c|}{\textbf{Long-context}} & \multicolumn{2}{c}{\textbf{Short-context}} \\
 & & & \textbf{Short-form Answers} & \textbf{Free-form Answers} & \textbf{Short-form Answers} & \textbf{Free-form Answers}\\
\midrule
- & - & Gold Answer & $1.5$ & $45.1$ & $1.5$ & $45.1$ \\
\midrule
\multirow{10}{*}{Llama3.1} & \multirow{5}{*}{8B} & Direct QA & $120.3$ & $197.7$ & $109.3$ & $135.1$ \\
& & CoT & $35.8$ & $75.4$ & $23.3$ & $74.4$\\
& & PoT & $5.3$ & $26.6$ & $2.8$ & $25.1$\\
& & \ourmethod & $24.3$ & $54.8$ & $21.4$ & $53.8$\\
\cmidrule{2-7}
& \multirow{5}{*}{70B} & Direct QA  & $118.9$ & $220.1$ & $118.9$ & $220.1$ \\
& & CoT & $5.9$ & $44.2$ & $14.3$ & $46.2$ \\
& & PoT & $3.6$ & $30.0$ & $3.3$ & $36.4$ \\
& & \ourmethod & $17.1$ & $43.5$ & $16.5$ & $43.0$ \\
\midrule
\multirow{5}{*}{\texttt{gpt-4o}} & \multirow{5}{*}{-} & Direct QA & $151.2$ & $213.2$ & $105.7$ & $141.6$ \\
& & CoT & $30.1$ & $84.5$ & $20.7$ & $69.1$ \\
& & PoT & $3.0$ & $24.4$ & $4.6$ & $52.4$\\
& & \ourmethod & $10.6$ & $67.0$ & $10.6$ & $82.4$ \\
\bottomrule
\end{tabular}
\caption{Statistics of the number of tokens of gold answers and different results.}
\label{tab:token}
\end{table*}

In this section, we show the comparison of the number of tokens output by different methods and the number of tokens of gold answers.
(\emph{i})~It can be found that the number of tokens output by PoT is consistently lower than that of other methods, whether it is a short-form answer or a free-form answer, which explains to a certain extent the reason why PoT has low performance, especially on the free-form answers.
(\emph{ii})~On the contrary, the number of tokens output by the Direct QA is generally high, which also reveals the reason why its EM is $0$ on the short-form answers.
(\emph{iii})~And \ourmethod is the closest in quantity to the number of tokens of gold answer, which shows that \ourmethod can adapt to obtain answers of various reasoning types.
    
\end{document}